\documentclass{article}

\usepackage[final]{corl_2026}

\usepackage{amsmath,amssymb,amsfonts}
\usepackage{algorithmic}
\usepackage{graphicx}
\usepackage{textcomp}

\newcommand{\jointp}{\mathbf{q}}
\newcommand{\dq}{\dot{\mathbf{q}}}
\newcommand{\ddq}{\ddot{\mathbf{q}}}
\newcommand{\torque}{\boldsymbol{\tau}}
\newcommand{\torqued}{\boldsymbol{\tau}_d}
\newcommand{\torqueext}{\boldsymbol{\tau}_\text{ext}}
\newcommand{\fext}{\mathbf{f}_\text{ext}}
\newcommand{\massmatrix}{\mathbf{M}}
\newcommand{\cartmassmatrix}{\Lambda}
\newcommand{\coriolis}{\mathbf{C}}
\newcommand{\gravity}{\mathbf{g}}
\newcommand{\jacobian}{\mathbf{J}}
\newcommand{\stiffnessx}{\mathbf{K}_x}
\newcommand{\dampingx}{\mathbf{D}_x}
\newcommand{\error}{\Tilde{\mathbf{x}}}
\newcommand{\derror}{\Tilde{\dot{\mathbf{x}}}}
\newcommand{\pose}{\mathbf{T}}

\newcommand{\dx}{\dot{\mathbf{x}}}
\newcommand{\ddx}{\ddot{\mathbf{x}}}
\newcommand{\dddx}{\dddot{\mathbf{x}}}

\newcommand{\ourwork}{PAKT}

\newcommand{\al}{&}

\providecommand{\appendices}{\appendix}

\title{\ourwork{}: Physically-Aligned Kinesthetic Teaching for Reinforcement Learning}

\author{
  Lars Johannsmeier\\
  NVIDIA\\
  \texttt{ljohannsmeie@nvidia.com}
  \And
  Yashraj Narang\\
  NVIDIA\\
  \texttt{ynarang@nvidia.com}
}

\hypersetup{
  pdftitle={PAKT: Physically-Aligned Kinesthetic Teaching for Reinforcement Learning},
  pdfauthor={Lars Johannsmeier and Yashraj Narang}
}

\makeatletter
\renewcommand{\@conferencelocation}{Austin, Texas, USA}
\makeatother

\AtBeginDocument{%
  \selectfont
}

\begin{document}
\maketitle

\begin{abstract}
Real-world reinforcement learning (RL) systems still struggle with the demands of contact-rich industrial manipulation, including micrometer-level precision, success rates above $99$\%, and human-level cycle times. Although off-policy algorithms can improve performance by leveraging demonstrations and interventions, a key bottleneck is the lack of an intuitive interface for collecting such guidance while complying with constraints of the physical system and the policy. We propose \ourwork{}, a framework for kinesthetic teaching in real-world RL. As opposed to teleoperation approaches, \ourwork{} relies on kinesthetic guidance, which is widely used in industry. However, a critical weakness of kinesthetic guidance is the possibility for the operator to move the robot along trajectories (e.g., velocities, accelerations, jerk) that the robot and/or policy cannot physically reproduce. Using \ourwork{}, operators guide the robot through admittance control, which maps human-applied forces to motion. The downstream reference generator applies the same kinematic limits used during policy execution, keeping the collected trajectories within these limits. To support this teaching interface with an appropriate execution layer, \ourwork{} adds a high-performance control stack that maps low-frequency RL actions to high-frequency torque commands. It consists of a reference generator and subsequent impedance controller, where the reference generator preserves the tracking performance of the impedance controller while improving contact handling and producing smoother policy actions. Across the reported runs on four insertion and industrial assembly benchmarks, including a data center compute tray, the end-to-end system reduces cycle time by $23$\%--$48$\% and cumulative intervention count by $62$\%--$86$\% relative to the HIL-SERL baseline. The hybrid ablation indicates that the controller stack accounts for most of the cycle-time reduction. With the controller held fixed, kinesthetic guidance reduces initial demonstration time by $8$\%--$22$\% and cumulative intervention count by $34$\%--$64$\%, while final success and cycle time remain similar. RAM insertion is evaluated across five random seeds, while the other tasks are single runs.
Project website: \href{https://pakt-website.github.io/pakt-website}{https://pakt-website.github.io/pakt-website}
\end{abstract}


\section{Introduction}
High-precision, contact-rich industrial manipulation remains difficult to automate with current robot learning systems.
Tasks such as electronics assembly require tight tolerances, high success rates, short cycle times, and reliable behavior under contact.
Reinforcement learning (RL) is attractive because it can adapt directly from real-world interaction, but practical deployment is still limited by sample efficiency \cite{tang2025deep}, robustness requirements, and the difficulty of transferring precise contact-rich behavior from simulation to hardware \cite{blanco2024benchmarking,tang2025deep,aljalbout2026reality,hofer2021sim2real,zhao2020sim}.
As a result, real-world demonstrations and human interventions remain central to many successful systems \cite{luo2025precise,tang2025deep}.

A key bottleneck is the interface through which this human guidance is collected. Teleoperation devices can provide useful demonstrations, e.g., for pick-and-place-style tasks \cite{dass2024telemoma}, but they often impose unintuitive mappings, require operator training \cite{chen2025teleoplab}, or add dedicated hardware \cite{zhao2024aloha,fu2024mobile,wu2024gello}.
Kinesthetic teaching allows the operator to guide the robot directly and is widely used in industry.
However, naive kinesthetic teaching introduces a critical mismatch for reinforcement learning: the human may move the robot along trajectories with velocities, accelerations, or jerks that the learned policy and downstream controller cannot reproduce.
Demonstrations that are easy for a human to provide are therefore not necessarily feasible for the policy to imitate \cite{chi2024universal}.

\begin{figure*}[ht!]
    \centering
    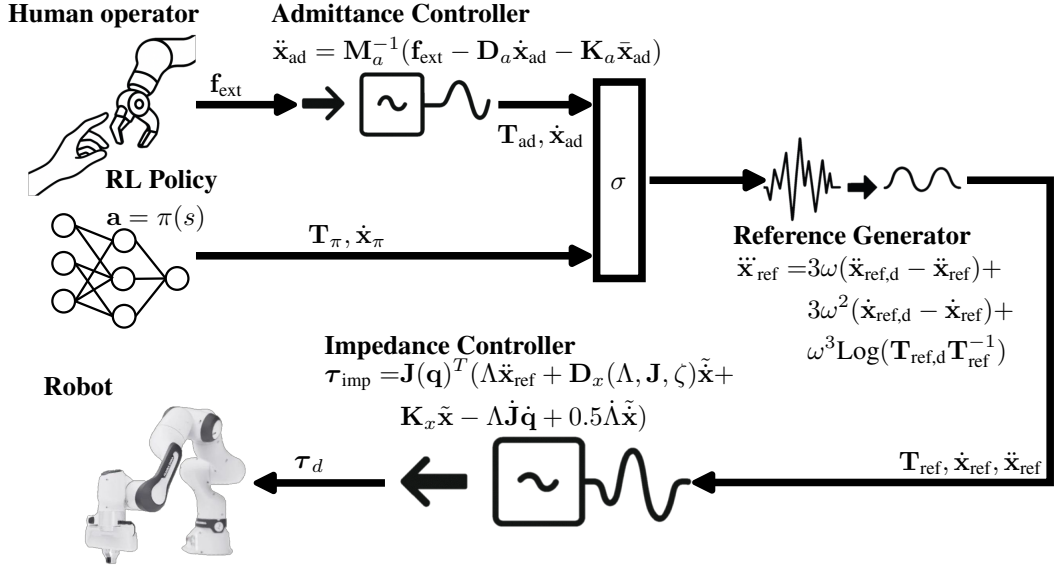
    \caption{Overview of \ourwork{}. In policy mode, policy actions are sent directly to the reference generator, whereas during guiding mode, human-applied forces first pass through an admittance controller. The reference generator outputs a filtered, limit-compliant command for the impedance controller, which computes the final torque command. The policy runs at $10$~Hz and the downstream controller stack at $1000$~Hz.}
    \label{fig:overview}
\end{figure*}

We introduce \ourwork{}, an integrated teaching and execution interface for real-world RL (Fig.~\ref{fig:overview}). Its admittance, reference-generation, and impedance components are established methods. The contribution is their coupling to HIL-SERL \cite{luo2025precise} so that human corrections and policy actions share the same 6-DoF coordinates, kinematic limits, reference dynamics, and tracker. This keeps stored guide actions inside the policy action space and executes them through the same bounded dynamics used during autonomous policy execution.

To support this interface, \ourwork{} couples the teaching layer with a high-performance control stack.
Low-frequency RL actions are passed through a reference generator that enforces the same limits as the admittance controller before being tracked by a fully compensated Cartesian impedance controller running at high frequency.
This design preserves accurate tracking, improves contact handling, and smooths policy execution without changing the high-level learning algorithm.

We evaluate \ourwork{} on four contact-rich insertion and industrial assembly tasks, including RAM insertion and data center compute-tray assembly. Relative to the HIL-SERL baseline \cite{luo2025precise}, which uses a SpaceMouse interface \cite{smith2024augmented}, the end-to-end system reduces cycle time by $23$\%--$48$\%, cumulative intervention count by $62$\%--$86$\%, and initial demonstration time by $15$\%--$34$\%. The baseline-to-hybrid comparison indicates that the controller stack accounts for most of the cycle-time reduction. Under the shared controller, kinesthetic guidance reduces initial demonstration time by $8$\%--$22$\% and cumulative intervention count by $34$\%--$64$\% relative to the SpaceMouse, while final success and cycle time remain similar. These measurements come from one trained operator. Only RAM insertion is repeated across five random seeds.

\section{Related Work}
Our framework sits at the intersection of real-world robotic RL, smooth motion generation, and human-in-the-loop learning \cite{han2023survey,tang2025deep,retzlaff2024human,mosqueira2023human}.
Controller design is particularly relevant for contact-rich robot learning.
Many successful real-world systems are deployed on position-controlled or otherwise simplified low-level interfaces \cite{han2023survey,tang2025deep,universalrobots2025urseries}, but contact-rich tasks often benefit from torque-controlled platforms and stronger downstream control stacks \cite{haddadin2024franka,campeau2019kinova,hirzinger2002dlr,dietrich2015overview}.
For physical interaction, impedance control \cite{hogan1985impedance,ott2008passivity} and admittance control \cite{keemink2018admittance} remain standard tools.
In RL, however, the downstream controller is often reduced to the simplest approach: pose-error and velocity feedback with manually tuned stiffness and damping.
We instead integrate a controller with desired-motion tracking, dynamics compensation, and automatic damping design \cite{albu2003cartesian}.
These ensure that the robot can accurately follow a desired motion and is always optimally damped in Cartesian space.

A second related line of work addresses the mismatch between policy outputs and smooth robot motion.
RL actions are often jerky \cite{raffin2022smooth,yang2025beyond}, motivating remedies such as smoothness-aware exploration and action regularization \cite{raffin2022smooth,raffin2020generalized}.
By contrast, classical robotics routinely relies on jerk-bounded or online trajectory generation to satisfy kinematic limits while preserving execution quality \cite{lee2024performance}.
Recent work has started to combine RL with jerk-bounded trajectory generators explicitly \cite{kolagar2025combining}.
Our reference generator follows the same overall motivation, but is intentionally lightweight and integrated directly upstream of the impedance controller.

A third line of work studies how human assistance can improve data collection and policy correction.
Human intervention can substantially accelerate real-world RL training \cite{luo2025precise}, and existing interfaces span 3D mice, modular teleoperation systems, and other operator-facing control devices \cite{dhat2024using,dass2024telemoma,smith2024augmented}.
Concurrent with our work, CR-DAgger uses compliance control to collect on-policy human corrections and trains a residual policy on top of a pretrained base policy \cite{xu2025crdagger}.
In contrast, \ourwork{} targets real-world RL directly and couples policy-aligned kinesthetic teaching with a shared downstream control stack rather than residual DAgger-style policy correction.
An important difference is that \ourwork{} pauses the policy and allows the operator to perform fine-grained interventions while CR-DAgger requires them to act concurrently with the active policy.
More recent systems also explore vision-based or low-cost leader-follower teleoperation and demonstration interfaces \cite{qin2023anyteleop,wu2024gello,fu2024mobile,chi2024universal}.
Kinesthetic teaching has likewise been studied directly as a user-facing programming interface for redundant manipulators \cite{wrede2013user}.
We build on this literature by using admittance control \cite{keemink2018admittance} to translate human-applied forces into motion while keeping demonstrations inside the same kinematic limits and action parameterization seen by the policy and downstream control stack.
Direct-input alternatives include GELLO \cite{wu2024gello}, which provides pose commands through a replica arm and avoids camera occlusion, but requires leader hardware and calibration. Both approaches still depend on downstream action bounds. Dall'Alba and Boriero \cite{dallalba2025intuitive} compared gamepad teleoperation with kinesthetic teaching in a 20-participant industrial programming study. The gamepad produced shorter trajectories, fewer waypoints, and lower interaction forces, while requiring more programming time. Their results show that interface comparisons depend on the task, operator population, and evaluation metric. Our experiments instead study demonstrations and interventions during online RL under a shared constrained controller, but do not establish a population-level interface preference.

\section{Methods}
\subsection{Reinforcement Learning}\label{sec:methods:learning}
Each manipulation task is modeled as a continuous-control Markov Decision Process (MDP) $\mathcal{M}= \{\mathcal{S},\mathcal{A},\rho,\mathcal{P},r,\gamma\}$
where $s_t \in \mathcal{S}$ denotes the state, $a_t \in \mathcal{A}$ the action, $\rho$ the initial state distribution, $\mathcal{P}(s_{t+1}|s_t,a_t)$ the transition dynamics, and $r_t(s_t,a_t)$ the task reward.
The policy does not receive the full state $s_t$. It acts on an observation $o_t \in \mathcal{O}$ and selects actions according to $\pi_\theta(a_t|o_t)$.
The parameterized policy induces trajectories $\tau=(s_0,o_0,a_0,\dots,s_H,o_H,a_H)$ and aims to maximize the expected discounted return over horizon $H$ as
$J(\pi_\theta) = \mathbb{E}_{\tau \sim \pi_\theta} \left[ \sum_{t=0}^{H} \gamma^t r(s_t, a_t) \right]$.

The observation $o_t$ comprises two RGB wrist-camera images, Cartesian pose with orientation represented by zyx Euler angles, and Cartesian velocity.
Actions consist of $6$-DoF Cartesian delta poses.
Similar to \cite{luo2025precise}, commands are provided to the control stack at $10$ Hz.
We keep this update rate for direct comparability with the baseline.

We use reinforcement learning with prior data (RLPD) \cite{ball2023efficient} to find the optimal policy.
RLPD extends Soft Actor-Critic (SAC) by combining two replay buffers $\mathcal{D}_\text{demo}$ and $\mathcal{D}_\text{rl}$. $\mathcal{D}_\text{demo}$ contains human demonstration data and $\mathcal{D}_\text{rl}$ contains experience collected by the robot during training.
At each iteration, a minibatch $B \subseteq \mathcal{D}_\text{demo} \cup \mathcal{D}_\text{rl}$ is sampled to update the critic and actor:

\begin{align}
L_Q &= \mathbb{E}_{(o_t,a_t,r_t,o_{t+1})\sim B} \Big[ \big(Q_\phi(o_t,a_t) - [r_t + \gamma \left(\min_i Q_{\bar{\phi}_i}(o_{t+1},a_{t+1}) - \alpha \log \pi_\theta(a_{t+1}|o_{t+1})\right)]\big)^2 \Big], 
\\
L_\pi &= \mathbb{E}_{\substack{o_t\sim B,\\ a_t \sim \pi_\theta(\cdot|o_t)}} \Big[ \alpha \log \pi_\theta(a_t|o_t) - Q_\phi(o_t,a_t) \Big],
\end{align}

where $L_Q$ is the critic loss, $L_\pi$ the actor loss, $Q_\phi$ the soft Q-function, $Q_{\bar{\phi}_i}$ the target critic network, $a_{t+1} \sim \pi_\theta(\cdot|o_{t+1})$, and $\alpha$ is an entropy temperature automatically tuned to maintain a target policy entropy.
Sampling equally from $\mathcal{D}_\text{demo}$ and $\mathcal{D}_\text{rl}$ ensures stable bootstrapping from human data and progressive adaptation to self-collected experience.

\subsubsection{Human-in-the-Loop Corrections}

A human operator supervises execution and can intervene via a 6-DoF input device or through kinesthetic guidance whenever unsafe or suboptimal behavior arises.
During RL, a held-button dead-man pauses the policy and enables guide mode. On entry, the reference and admittance poses are reset to the measured pose with zero derivatives. This reduces command discontinuity but does not provide a formal worst-case force guarantee.
During an intervention, the policy is overridden by either SpaceMouse input or kinesthetic guidance, according to the experimental condition. All resulting observation-action-reward tuples are logged as corrective data.
Once intervention ends, these trajectories are added to both $\mathcal{D}_\text{demo}$ and $\mathcal{D}_\text{rl}$ for subsequent training.
This mechanism allows continuous refinement without resetting the learning process.

\subsection{Reward and Success Classification}

We use a binary reward derived from a learned success classifier $f_\psi(o_t)$ and a sigmoid function so that $r_t=1$ for $\text{sigmoid}(f_\psi(o_t)) > 0.85$ and $0$ otherwise.
More information on this can be found in Appendix \ref{app:classifier}.

\subsection{Controller}\label{sec:methods:controller}
The input to our control stack is either (1) a desired Cartesian delta pose $\pose_\text{a}$ from the RL policy or (2) an external wrench $\fext$ induced by the human operator if guide mode is activated.
In Case 1, the input goes directly to a reference generator, and in Case 2, the input goes to an admittance controller first.
The output of the admittance controller is then passed to the reference generator.
Next, the output of the reference generator is forwarded to the impedance controller, which calculates torque commands for the robot.
All three blocks execute at $1$~kHz and have distinct roles. The admittance controller maps a human-applied wrench to motion, the reference generator applies kinematic bounds to human or policy targets, and the impedance controller tracks the resulting reference.
Figure~\ref{fig:overview} provides a visualization of the stack.
Throughout this section, we omit explicit dependence on $t$ for clarity.

$\pose_\text{a}$ is composed with the current pose of the end effector as $\pose_\pi = \pose_\text{EE}\pose_\text{a}$
where $\pose_\pi$ is the new desired pose expressed in world frame, $\pose_\text{EE}$ is the current end effector pose expressed in world frame, and the policy action $\pose_\text{a}$ is expressed in body frame.
We estimate the desired twist of the policy as $\dx_\pi = k_v \text{Log}( \pose_\pi\pose_\text{EE}^{-1})$
where $\text{Log}$ is the logarithmic map and $k_v=\frac{1}{\Delta t}$ is a proportional factor.
We set $\ddx_\pi=\boldsymbol{0}$ in our implementation because its estimate is too noisy.

When guiding mode is active, the admittance controller provides the desired pose, twist, and acceleration to the reference generator instead of the policy.
The admittance controller is defined as
\begin{equation}
    \ddx_\text{ad} = \mathbf{M}_a^{-1}(\fext-\mathbf{D}_a \dx_\text{ad} - \mathbf{K}_a \error_\text{ad})
\end{equation}
where $M_a$ is the virtual mass matrix, $D_a$ the virtual damping matrix, $K_a$ is the virtual stiffness matrix, and $\error_\text{ad}=\text{Log}(\pose_\text{ad}\pose_\text{EE}^{-1})$ is the pose error.
We set $K_a=\boldsymbol{0}$, allowing the robot to be guided freely through Cartesian space.

Then, we retrieve the setpoint for the reference generator:
\begin{equation}
    \dx_\text{ad}=\int \ddx_\text{ad} dt,\quad \pose_\text{ad}=\exp (\hat{\mathbf{V}}\Delta t)\pose^\text{prev}_\text{ad},\quad \hat{\mathbf{V}}=\left[\begin{array}{cc} [\boldsymbol{\omega}]_x& \boldsymbol{v} \\ \boldsymbol{0}& \boldsymbol{0} \end{array}\right]
\end{equation}
where $\dx_\text{ad}$ is the integrated twist, $\pose_\text{ad}$ the pose (where the superscript prev indicates the pose from the previous step), $[\boldsymbol{\omega}]_x$ is the skew-symmetric matrix, and $\boldsymbol{v}$ the linear velocity.

The reference generator is a critically-damped, third-order filter with the general dynamics
\begin{equation}
    \dddx_\text{ref}=a_2 (\ddx_\text{ref,d}- \ddx_\text{ref}) + a_1 (\dx_\text{ref,d}- \dx_\text{ref}) + a_0 \text{Log}(\pose_\text{ref,d}\pose_\text{ref}^{-1})
\end{equation}

where
\begin{equation}
\left(\pose_\text{ref,d},\dx_\text{ref,d},\ddx_\text{ref,d}\right)=
\begin{cases}
\left(\pose_\text{ad},\dx_\text{ad},\ddx_\text{ad}\right), & \sigma = 1,\\
\left(\pose_\pi,\dx_\pi,\ddx_\pi\right), & \sigma = 0,
\end{cases}
\end{equation}
Here, $\pose_\text{ref,d}$, $\dx_\text{ref,d}$, and $\ddx_\text{ref,d}$ denote the desired pose, twist, and acceleration supplied to the reference generator. The binary variable $\sigma \in \{0,1\}$ indicates the control mode: $\sigma=1$ activates guiding mode, whereas $\sigma=0$ means that the policy drives the robot.

We choose a triple pole at $-\omega$, the bandwidth, which yields the critically-damped system

\begin{equation}
\dddx_\text{ref}=3\omega (\ddx_\text{ref,d}- \ddx_\text{ref}) + 3\omega^2 (\dx_\text{ref,d}- \dx_\text{ref}) + \omega^3 \text{Log}(\pose_\text{ref,d}\pose_\text{ref}^{-1}).
\end{equation}

$\dddx_\text{ref}$ is integrated three times to obtain $\pose_\text{ref}$ while we apply user-provided limits $\dddx_\text{lim},\ddx_\text{lim},\dx_\text{lim}$ at each integration step:
\begin{equation}
    \dddx_\text{ref}=\text{clip}(\dddx_\text{ref},\dddx_\text{lim}),\quad \ddx_\text{ref}=\text{clip}(\ddx_\text{ref},\ddx_\text{lim}),\quad \dx_\text{ref}=\text{clip}(\dx_\text{ref},\dx_\text{lim})
\end{equation}

All three control blocks continue to execute at $1$~kHz during guidance. Guide trajectories are downsampled and logged at the $10$-Hz policy rate. At each logged step, we store the normalized body-frame delta $\pose_{\mathrm{a},t}=\pose_{\mathrm{EE},t}^{-1}\pose_{\mathrm{ad},t}$ after applying the same limits used for execution. Limiting the action before storage ensures that the logged action generated the corresponding transition. Post-hoc clipping would instead associate a changed action with a transition produced by another command. We did not log the frequency with which the limits saturated.

The reference generator's output is then passed to the Cartesian impedance controller.
Our controller stack assumes the robot's dynamics as
\begin{equation}
    \massmatrix(\jointp)\ddq + \coriolis(\jointp,\dq)\dq + \gravity(\jointp)=\torqued + \torqueext,
\end{equation}
where $\massmatrix$ is the mass matrix, $\coriolis$ is the Coriolis matrix, $\gravity$ is the gravity torque vector, $\jointp, \dq, \ddq$ are joint position, velocity, and acceleration, $\torqued$ is the desired torque, and $\torqueext$ is the measured external torque.
We use an impedance controller with the equation
\begin{equation}\label{eq:impedance_controller}
    \torque_\text{imp}=\jacobian(\jointp)^T\left[\cartmassmatrix \ddot{\mathbf{x}}_\text{ref}+ \dampingx(\cartmassmatrix,\stiffnessx, \mathbf{\zeta}) \derror + \stiffnessx \error - \cartmassmatrix \dot{\jacobian}(\dq) \dq + 0.5 \dot{\cartmassmatrix} \derror\right],
\end{equation}
where $\cartmassmatrix$ is the Cartesian mass matrix, $\jacobian$ is the Jacobian, $\stiffnessx$ is the Cartesian stiffness matrix, $\dampingx$ is the Cartesian damping matrix with damping design based on double-diagonalization according to \cite{albu2003cartesian} with damping factors $\mathbf{\zeta}$, and $\error=\text{Log}(\pose_\text{ref}\pose_\text{EE}^{-1})$ is the Cartesian space pose error.
Double diagonalization transforms the Cartesian mass and stiffness matrices into modal space, where the damping matrix can be designed as a diagonal matrix and then transformed back into Cartesian space.
The term $\cartmassmatrix \dot{\jacobian} \dq$ compensates geometric acceleration, and $0.5 \dot{\cartmassmatrix}$ ensures passivity.
See \cite{albu2003cartesian} for a detailed discussion of this controller type.

We also use a nullspace controller to keep the elbow of our redundant robot upright.
It is defined as
\begin{equation}
    \torque_\text{null} = \mathbf{N}^\text{T} (\mathbf{D}_\text{null} \Tilde{\dq} + \mathbf{K}_\text{null} \Tilde{\jointp})
\end{equation}
where $\mathbf{D}_\text{null}$ is a damping matrix, $\mathbf{K}_\text{null}$ a stiffness matrix, $\Tilde{\jointp}=\jointp_\text{d,null}-\jointp$ and $\Tilde{\dq}=\dq_\text{d,null}-\dq$ denote the nullspace position and velocity errors, respectively, and $\mathbf{N} = \mathbf{I} - \massmatrix^{-1} \jacobian^T (\jacobian \massmatrix^{-1} \jacobian^T)^{-1} \jacobian$ is the dynamically consistent nullspace projector \cite{dietrich2015overview}.

Finally, the robot receives the resulting torque including compensation of Coriolis and gravity:
\begin{equation}
    \torqued=\torque_\text{imp}+\torque_\text{null} + \coriolis(\jointp,\dq)\dq + \gravity(\jointp)
\end{equation}

In Sec. \ref{sec:experiments:controller} we compare this controller stack to variants commonly used in the RL literature.
Table~\ref{tab:controller_equations} gives an overview with the respective equations and a comment that describes the main difference. C1 is stated directly at torque level, while the C2--C5 rows give Cartesian wrench terms with corresponding joint torque $\torque_\text{imp}=\jacobian(\jointp)^T\boldsymbol{f}_\text{imp}$.
C3 closely matches a controller structure that is common in the RL literature.

\begin{table}[ht!]
\caption{Variations of controllers used in RL. For C5, $\mathbf{e}_\mathrm{lim}$ denotes the clipped error. C5 is the controller used in our baseline \cite{luo2025precise}}
    \label{tab:controller_equations}
    \centering
    \resizebox{\columnwidth}{!}{%
    \begin{tabular}{|c|l|p{7.0cm}|}
    \hline
    \textbf{ID} & \textbf{Equation} & \textbf{Comment} \\
    \hline
    C1 & Eq. \ref{eq:impedance_controller} & Ours \\
    \hline
        C2 & $\boldsymbol{f}_\text{imp}=-\dampingx(\cartmassmatrix,\stiffnessx,\mathbf{\zeta}) \dx + \stiffnessx \error$ & No inertia compensation or desired velocity \\
        \hline
         C3 & $\boldsymbol{f}_\text{imp}=-\dampingx \dx + \stiffnessx \error$ & No inertia compensation or desired velocity, and no damping design \\
        \hline
        C4 & $\boldsymbol{f}_\text{imp}=-\dampingx \dx + \stiffnessx \error + \int k_i \error dt$ & No inertia compensation or desired velocity, and no damping design, but with integrator \\
        \hline
        C5 & $\boldsymbol{f}_\text{imp}=-\dampingx \dx + \stiffnessx \mathrm{clip}(\error, \mathbf{e}_\mathrm{lim})$ & No inertia compensation or desired velocity, and no damping design, but with error clipping \\
         \hline
    \end{tabular}
    }
\end{table}

\section{Experiments}
\subsection{Setup}\label{sec:experiments:setup}
Our experimental setup consists of a Franka Emika robot \cite{haddadin2024franka} equipped with a Franka Hand end effector.
The complete controller stack runs on an NVIDIA Jetson Orin AGX with a real-time kernel, while the policy and success classifier run on a desktop PC with an NVIDIA RTX 4090 GPU.
Attached to the robot end effector are two Intel RealSense D405 cameras.

\subsection{Controller Stack Comparison}\label{sec:experiments:controller}
Our first set of experiments compares the proposed control stack and the controller variants listed in Tab.~\ref{tab:controller_equations} in terms of tracking and contact response.
Cartesian pose commands are issued at $10$~Hz, matching the RL experiments.
For tracking, the robot follows a square in the xy-plane with sinusoidal motion in z.
For contact, it moves toward a hard surface in the z-direction until the observed external force exceeds $10$~N.

\begin{figure*}[ht!]
    \centering
    \includegraphics[width=\columnwidth]{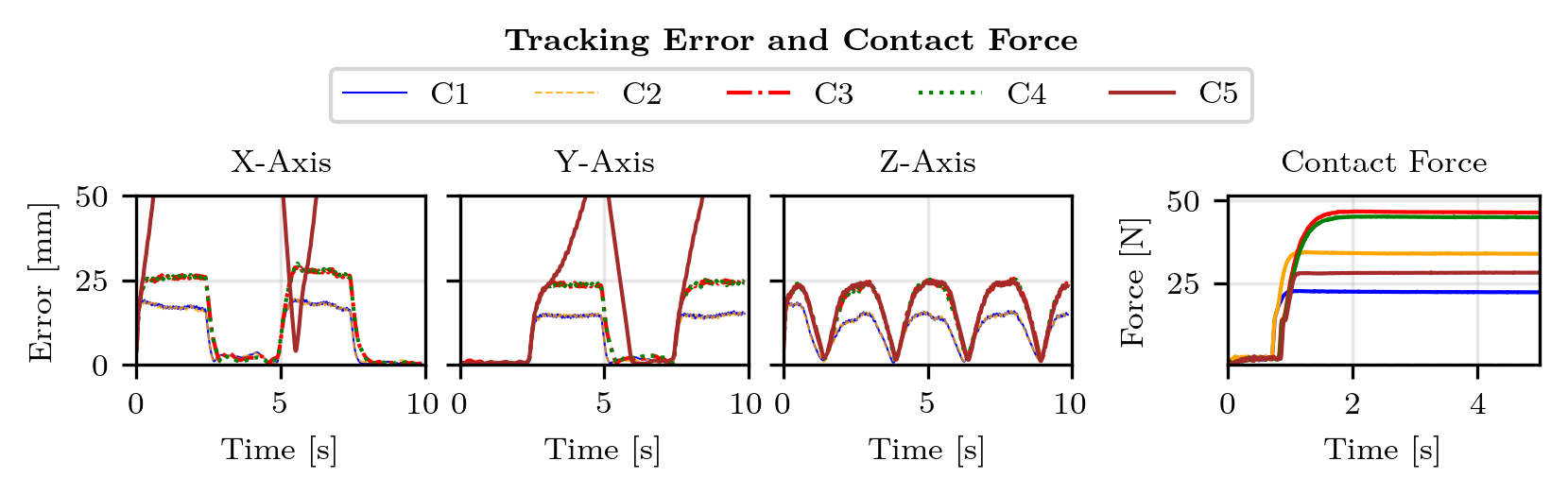}
    \caption{Tracking error for the tested control stacks in the x, y, and z dimensions. Tracking errors are low-pass filtered at $7$~Hz for visibility. We truncate C5 to maintain visibility for the other time series, since error clipping makes C5 lag behind the reference trajectory and therefore produce much larger absolute tracking errors than the other stacks. C1 and C2 overlap in the tracking error plots, as do C3 and C4.}
    \label{fig:controller_comparison}
\end{figure*}

The left three plots of Fig.~\ref{fig:controller_comparison} show that C2, the closest ablation of C1, achieves nearly identical tracking error.
This indicates that the upstream reference generator does not degrade the free-space tracking performance of the impedance controller.
By contrast, the simpler controller variants C3/C4 have substantially larger tracking error, with the integrator in C4 offering little benefit in this dynamic setting.
C5, the clipped-error baseline used in the RL comparison, performs worst because error clipping saturates the corrective action once the pose error exceeds the clipping threshold, causing large lag.
The right plot shows contact forces during the hard-contact test. The $10$~N stopping condition is evaluated at $10$~Hz to emulate governing-policy latency, while impedance control continues at $1$~kHz. The resulting overshoot therefore measures contact response under policy-rate supervision rather than a hard force bound. C1 and C2 have comparable tracking error, while C1 has a lower peak force ($22.5$~N versus $33.9$~N). C3/C4 have larger tracking errors and peak forces, and C5 reduces force through error clipping at the cost of severe tracking degradation. Table~\ref{tab:controller_comparison} reports the numeric values.

\begin{table*}[ht!]
\caption{Average tracking error and contact force for the tested controller stacks.}
    \label{tab:controller_comparison}
    \centering
    \begin{tabular}{|c|c|c|}
    \hline
    \textbf{Controller} & \textbf{Average tracking error [mm]} & \textbf{Contact force [N]} \\
    \hline
        C1 (\ourwork{}) & $\mathbf{\left[16.8,14.1,9.6\right]}$ & $\mathbf{22.5}$ \\
         \hline
         C2 & $\mathbf{\left[16.6,14.1,9.6\right]}$ & $33.9$ \\
         \hline
         C3 & $\left[24.8,22.5,16.2\right]$ & $46.5$ \\
         \hline
         C4 & $\left[24.8,22.4,16.0\right]$ & $45.0$ \\
         \hline
         C5 (HIL-SERL) & $\left[100.5,37.2,16.8\right]$ & $28.2$ \\
         \hline
    \end{tabular}
\end{table*}

\subsection{RL Experiments}\label{sec:experiments:learning}
Our second set of experiments compares \ourwork to the HIL-SERL baseline \cite{luo2025precise} and additionally to a hybrid variant that combines our controller stack with the baseline SpaceMouse interface.
This three-way comparison serves as a structured ablation: baseline versus hybrid isolates the controller stack, while hybrid versus \ourwork{} isolates kinesthetic guidance under the same controller.

\begin{figure*}[ht!]
\centering
\begin{tabular}{cccc}
\includegraphics[width=0.22\textwidth]{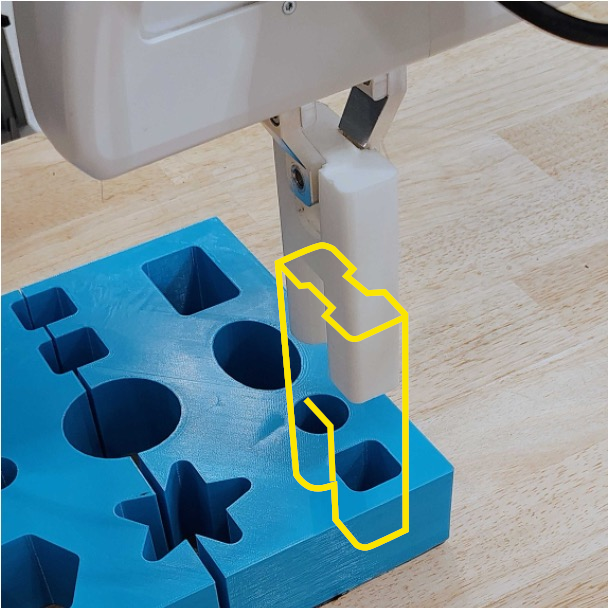} &
\includegraphics[width=0.22\textwidth]{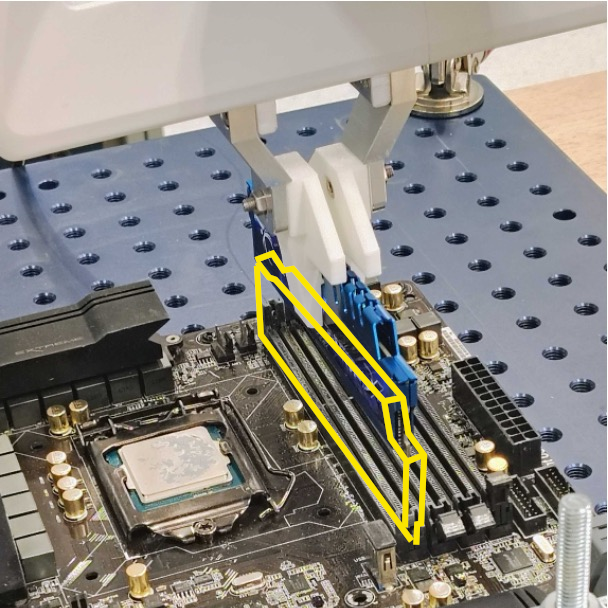} &
\includegraphics[width=0.22\textwidth]{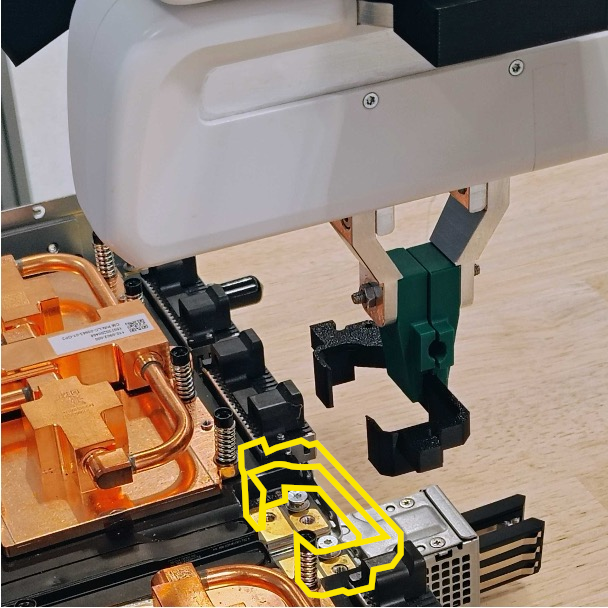} &
\includegraphics[width=0.22\textwidth]{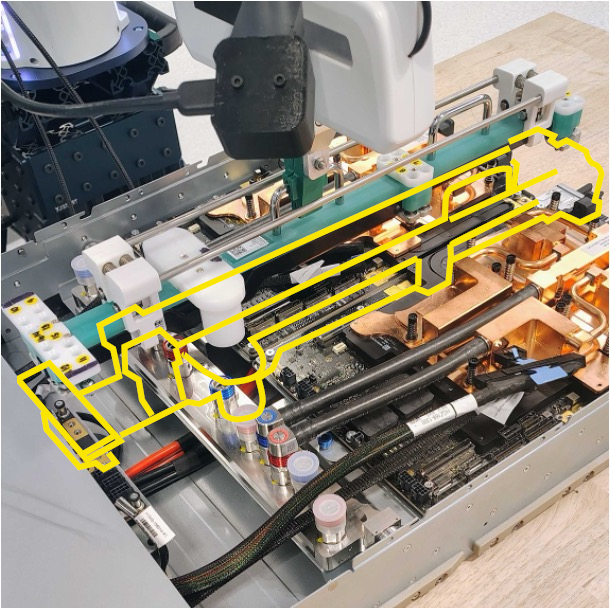} \\
\end{tabular}
\caption{The four insertion tasks used in the RL experiments with outlines indicating the insertion location: (a) FMB peg insertion, (b) RAM insertion, (c) limit fixture assembly, and (d) busbar assembly. We show a more detailed overview in Appendix \ref{app:setup} and in the supplementary video.}
\label{fig:setups}
\end{figure*}

Task details are given in Appendix \ref{app:setup}.
We consider four tasks: peg insertion from the Functional Manipulation Benchmark (FMB) \cite{luo2025fmb}, RAM insertion, limit fixture assembly, and busbar assembly.

All four tasks follow the same protocol. A single trained, unblinded operator performed all demonstrations and interventions. We first train a success classifier (Appendix \ref{app:classifier}), then collect $20$ demonstrations with either a SpaceMouse or kinesthetic guidance (Appendix \ref{app:demo}), train the policy with interventions, and finally evaluate the learned policy in $100$ executions from random initial states. The operator practiced with both interfaces until completing the tasks reliably. Practice rollouts were excluded from the reported times and replay buffers. During kinesthetic guidance, the operator's hands remained above the robot and outside both wrist-camera views. All variants use the same vision, pose, and velocity observations without a wrench channel. Within each task, the demonstration budget, reset protocol, and training procedure are held fixed. We repeated RAM insertion with five random seeds per method. The other tasks are single runs.

\begin{figure*}[ht!]
    \centering
    \includegraphics[width=\columnwidth]{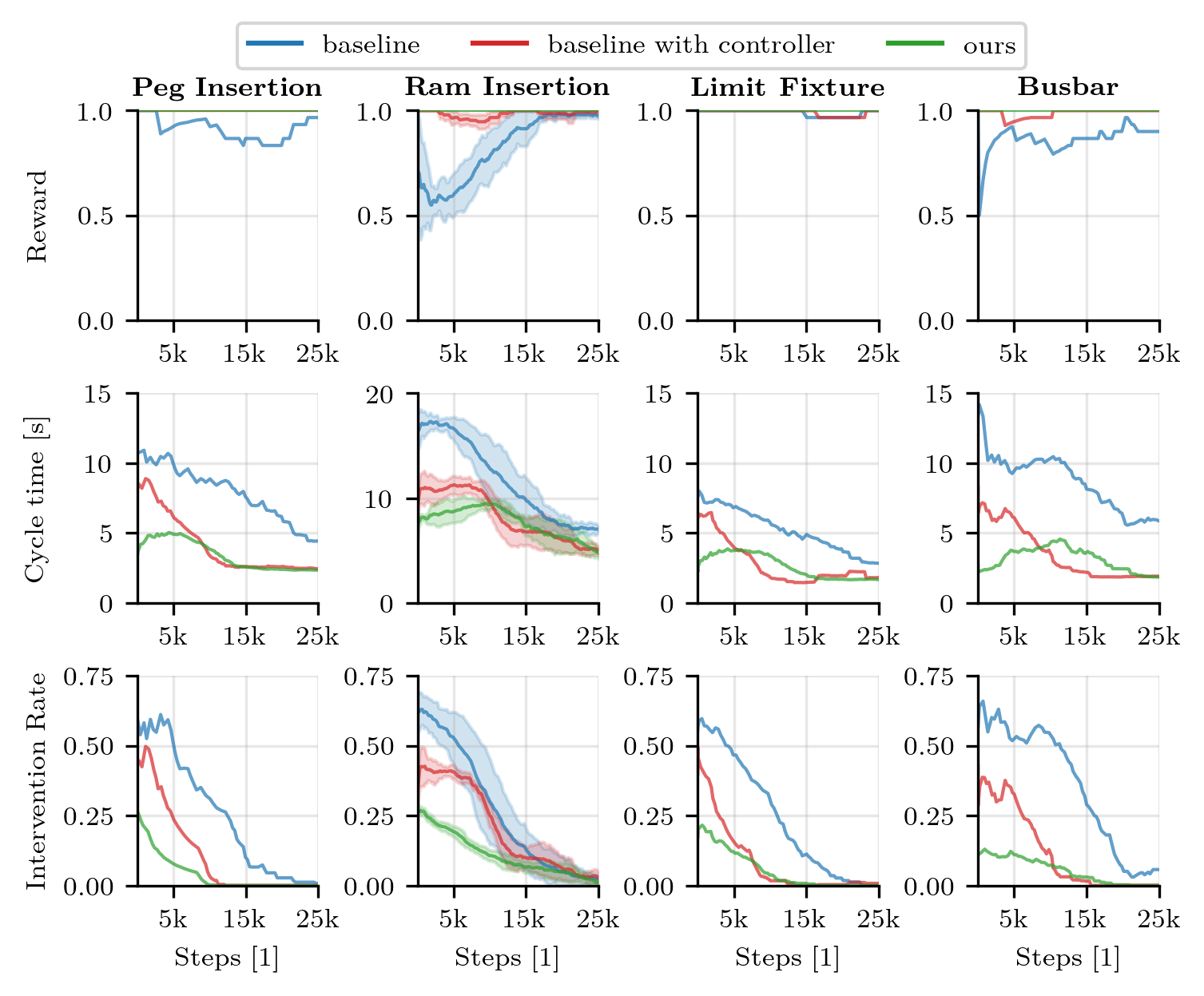}
    \caption{Reward, cycle time, and intervention rate over training for the baseline, hybrid, and \ourwork{}. RAM curves show the mean and standard deviation across five random seeds. The other tasks are single runs. \textbf{Top row:} The hybrid and \ourwork{} reach higher observed success rates than the baseline. \textbf{Middle row:} Their similar cycle times isolate the primary effect of the controller stack. \textbf{Bottom row:} The lower intervention rate of \ourwork{} relative to the hybrid isolates the effect of the kinesthetic interface.}
    \label{fig:results:rl}
\end{figure*}

Figure~\ref{fig:results:rl} shows reward, cycle time, and intervention rate over training, and Tab.~\ref{tab:results:rl} reports initial demonstration time, final success rate, average cycle time, and cumulative intervention count. For RAM insertion, the curves show the mean and standard deviation across five training runs with different random seeds, and the table reports the mean of each metric across the same five seeds. The other tasks are single runs. Relative to the baseline, both \ourwork{} and the hybrid reduce average cycle time by $23$\%--$48$\% across the reported runs. The similar cycle times of the hybrid and \ourwork{} indicate that the controller stack accounts for most of this reduction. Relative to the hybrid, kinesthetic guidance reduces initial demonstration time by $8$\%--$22$\% and cumulative intervention count by $34$\%--$64$\%, while final success and cycle time remain similar. Relative to the baseline, the end-to-end system reduces initial demonstration time by $15$\%--$34$\% and cumulative intervention count by $62$\%--$86$\%.
These measurements establish within-operator efficiency for the evaluated workflow, rather than population-level usability.

\begin{table}[ht!]
    \caption{Results on the four tasks for the HIL-SERL baseline (B), hybrid (H), and \ourwork{} (P). Cumulative intervention count is the discrete area under the intervention-rate curve over training steps. Average cycle time is computed over successful evaluation trials only. Failed and timed-out trials are excluded. Every entry in the RAM row is the mean across five seeds. The other rows are single runs.}
    \label{tab:results:rl}
    \centering
    \small
    \setlength{\tabcolsep}{2.3pt}
    \begin{tabular}{|l|ccc|ccc|ccc|ccc|}
    \hline
        \textbf{Task} & \multicolumn{3}{c|}{\textbf{\shortstack{Initial\\demonstration\\time [s]}}} & \multicolumn{3}{c|}{\textbf{Success rate}} & \multicolumn{3}{c|}{\textbf{Average cycle time [s]}} & \multicolumn{3}{c|}{\textbf{Cumulative interventions}}\\
        \cline{2-13}
        & \textbf{B} & \textbf{H} & \textbf{P} & \textbf{B} & \textbf{H} & \textbf{P} & \textbf{B} & \textbf{H} & \textbf{P} & \textbf{B} & \textbf{H} & \textbf{P}\\
        \hline
         Peg & $216$ & $199$ & $\mathbf{183}$ & $\mathbf{1.0}$ & $\mathbf{1.0}$ & $\mathbf{1.0}$ & $3.88$ & $\mathbf{2.25}$ & $\mathbf{2.17}$ & $6276$ & $2690$ & $\mathbf{980}$\\
         \hline
         RAM insertion & $243$ & $219$ & $\mathbf{182}$ & $0.88$ & $\mathbf{0.99}$ & $\mathbf{0.98}$ & $5.2$ & $\mathbf{4.07}$ & $\mathbf{4.0}$ & $8780$ & $5205$ & $\mathbf{3325}$\\
         \hline
         Limit fixture & $209$ & $194$ & $\mathbf{177}$ & $\mathbf{0.99}$ & $\mathbf{0.98}$ & $\mathbf{0.99}$ & $3.01$ & $\mathbf{1.87}$ & $\mathbf{1.92}$ & $5937$ & $2026$ & $\mathbf{1342}$\\
         \hline
         Busbar & $221$ & $187$ & $\mathbf{146}$ & $0.85$ & $\mathbf{0.98}$ & $\mathbf{0.99}$ & $3.91$ & $\mathbf{2.09}$ & $\mathbf{2.03}$ & $9286$ & $2920$ & $\mathbf{1327}$\\
         \hline
    \end{tabular}
    
\end{table}

\section{Limitations}
The interface comparison uses one trained, unblinded operator and reports initial demonstration time and intervention count. It therefore does not establish user-to-user generalization, usability, or a general advantage over teleoperation. In a 20-participant offline study, Dall'Alba and Boriero \cite{dallalba2025intuitive} found that gamepad teleoperation produced shorter trajectories and lower interaction forces but required more programming time. A counterbalanced multi-user study is needed. Kinesthetic guidance can also occlude external cameras, although the operator's hands remained outside the wrist-camera views in our experiments.

Our joint-torque-based wrench estimate cannot distinguish human-applied forces from environment contact during kinesthetic guidance. We use this estimate only for admittance control and do not include it in the policy observation. Joint friction can also bias the estimate and the resulting guidance dynamics. The quality of this estimate depends on the robot, and the Franka Emika Robot used here provides a high-quality external-wrench estimate. Bilateral teleoperation could provide a separate operator-command channel, but would not by itself guarantee separation of operator and environment effects.

The experiments cover single-arm insertion tasks on one Franka platform. Transfer to other robots and evaluations of non-insertion tasks, bimanual settings, and multiple operators remain untested. The same admittance setting was reused across all four tasks, but we did not conduct a sensitivity study or provide a stability or worst-case force guarantee. The evaluated impedance layer requires torque control. The admittance controller and reference generator could instead command a position-controlled robot, potentially extending the approach to a broader range of platforms, but this configuration remains untested.

\section{Conclusion}
\ourwork{} couples established admittance, reference-generation, and impedance-control components into a shared interface for policy actions and kinesthetic guidance. The reference generator applies velocity, acceleration, and jerk limits before a Cartesian impedance controller tracks the resulting motion. Human corrections are stored in the same 6-DoF action space and executed through the same bounded dynamics as policy actions.

Across the reported runs, the end-to-end system reduces cycle time by $23\%$--$48\%$, cumulative intervention count by $62\%$--$86\%$, and initial demonstration time by $15\%$--$34\%$ relative to HIL-SERL. The hybrid ablation indicates that the controller stack accounts for most of the cycle-time reduction. With the controller fixed, kinesthetic guidance reduces initial demonstration time by $8\%$--$22\%$ and cumulative intervention count by $34\%$--$64\%$, while final success and cycle time remain similar. These results are limited to one trained operator, with five-seed verification only for RAM insertion. A multi-operator study is required to assess workload, learnability, preference, and user-to-user generalization.

\bibliography{bibliography}

\clearpage
\appendices

\section{Experimental Setup}\label{app:setup}

This appendix summarizes the exact task configuration, learning setup, reward implementation, and data-collection protocol used in the real-world experiments.
Unless otherwise noted, all four tasks share the same hardware platform, policy-to-controller interface, and high-level training workflow.
Where task-specific settings differ, we report them explicitly in the tables below.

\begin{figure}[ht!]
    \centering
    \includegraphics[width=\columnwidth]{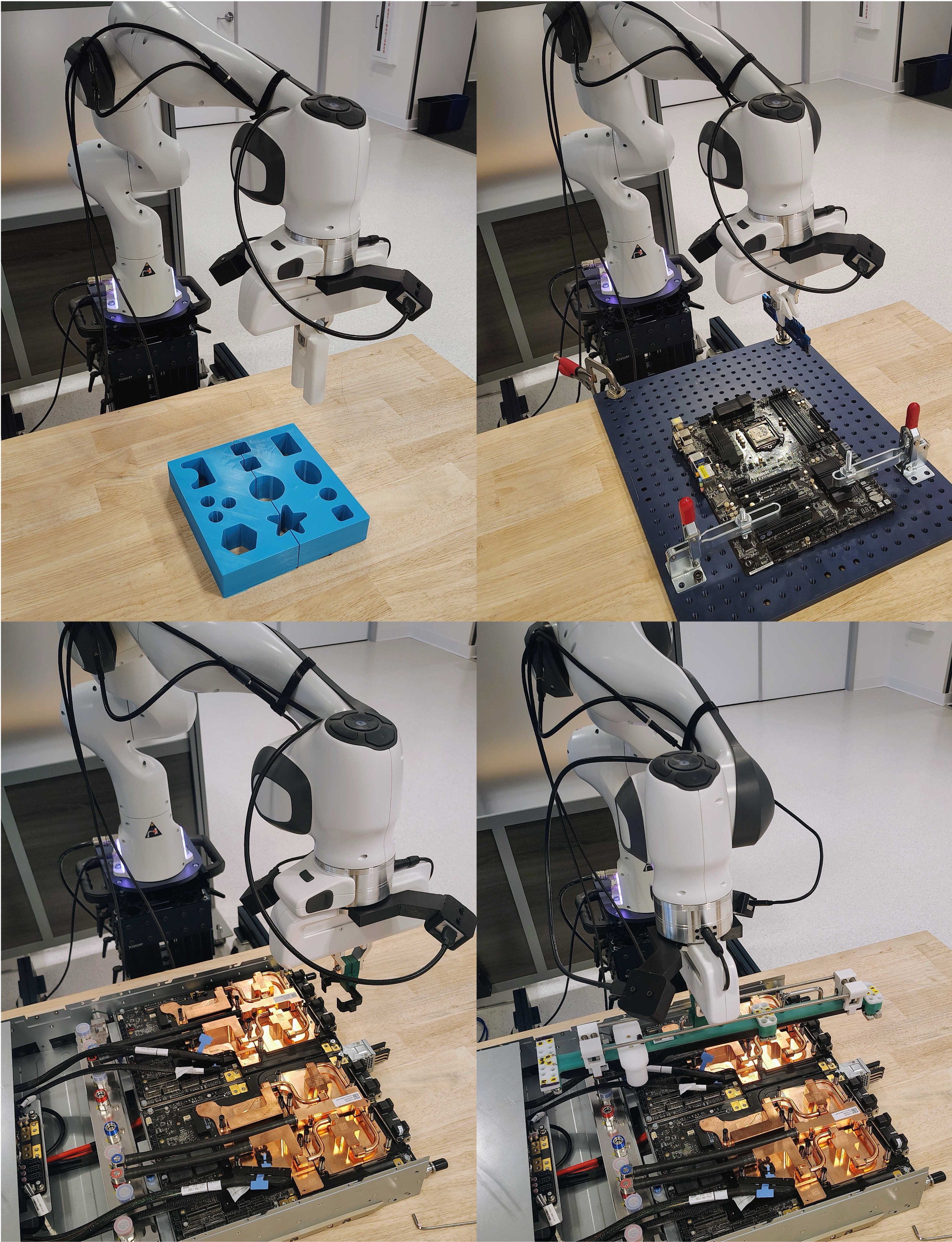}
    \caption{Experimental setup used for all real-world tasks.}
    \label{fig:experimental_setup}
\end{figure}

\begin{figure*}[ht!]
\centering
\begin{tabular}{cccc}
\includegraphics[width=0.22\textwidth]{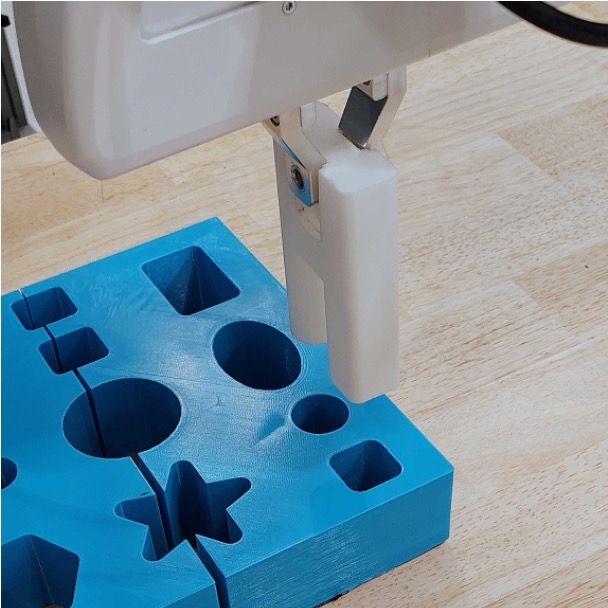} &
\includegraphics[width=0.22\textwidth]{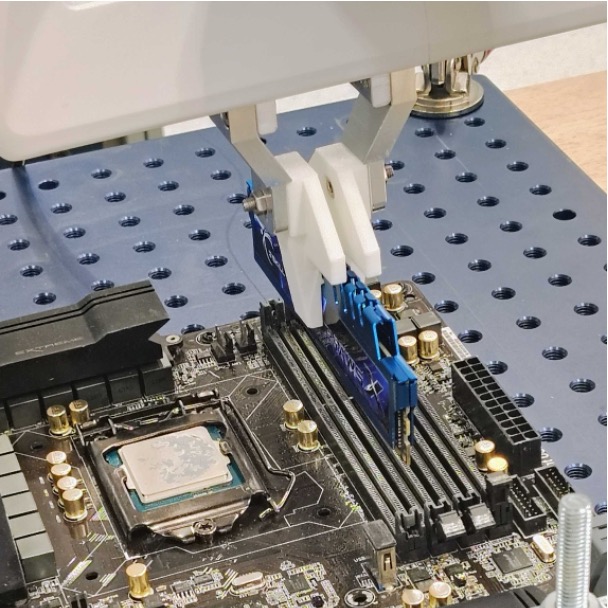} &
\includegraphics[width=0.22\textwidth]{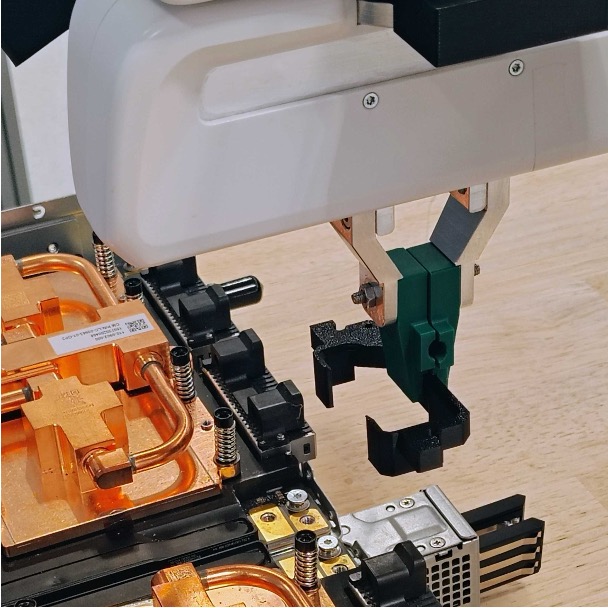} &
\includegraphics[width=0.22\textwidth]{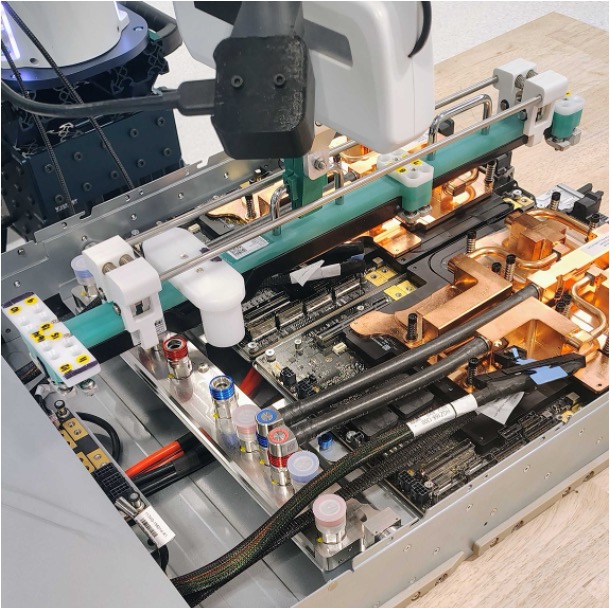} \\
\includegraphics[width=0.22\textwidth]{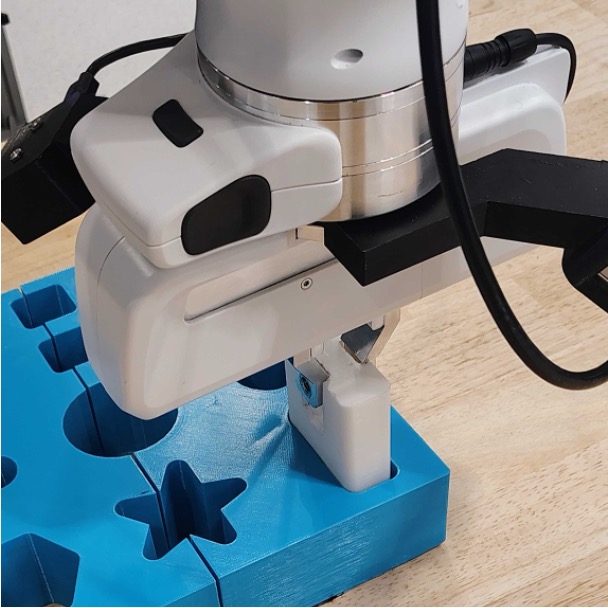} &
\includegraphics[width=0.22\textwidth]{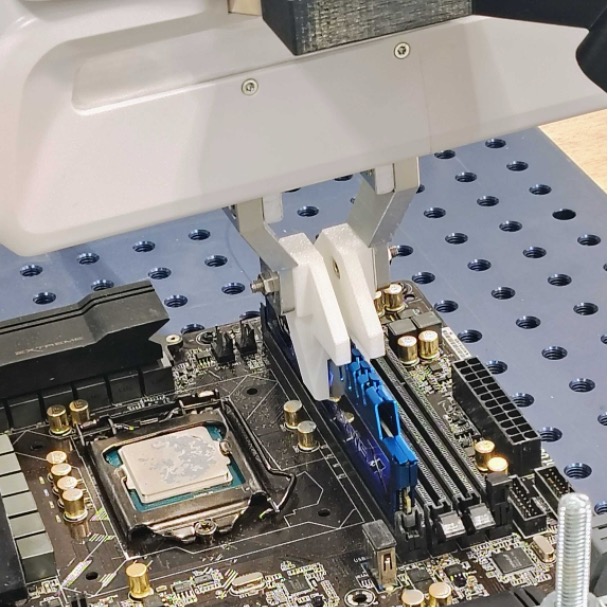} &
\includegraphics[width=0.22\textwidth]{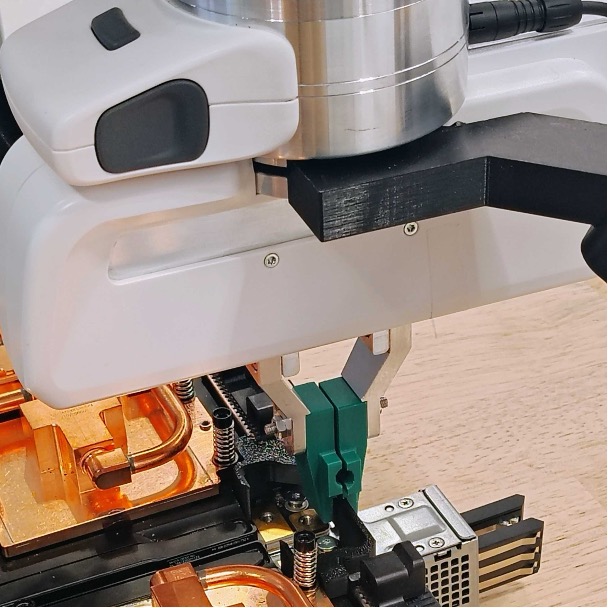} &
\includegraphics[width=0.22\textwidth]{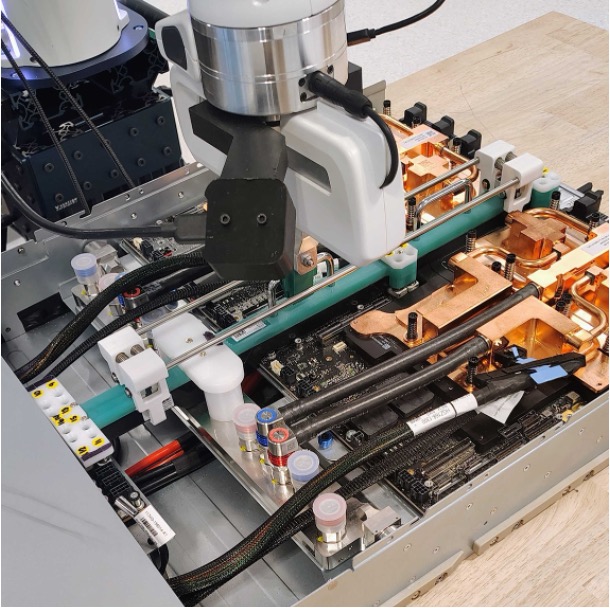} \\
\end{tabular}
\caption{The four insertion tasks used in the RL experiments: (a) FMB peg insertion, (b) RAM insertion, (c) limit fixture assembly, and (d) busbar assembly.}
\label{fig:setups_full}
\end{figure*}

Our experimental setup consists of a Franka Emika Robot \cite{haddadin2024franka} equipped with a Franka Hand end effector.
The complete controller stack runs on an NVIDIA Jetson Orin AGX with a real-time kernel, while the policy and success classifier run on a desktop PC with an NVIDIA RTX 4090 GPU.
Policy commands are sent at $10$~Hz to the controller stack, while the admittance controller, reference generator, and impedance controller execute in a $1$ kHz real-time loop on the Jetson.
Attached to the robot end effector are two Intel RealSense D405 cameras with the same custom mount as in \cite{luo2025precise}.
The Franka Hand is used only to establish and maintain the grasp for each episode. The studied tasks do not require regrasping or any form of in-hand manipulation.
Figure~\ref{fig:experimental_setup} shows the full setup, and Fig.~\ref{fig:setups_full} shows closeups of the respective tasks.

\subsection{Shared Setup}

Table~\ref{tab:appendix:shared_setup} lists the setup details shared across all tasks.
The policy observation and action interfaces are also shared across tasks. The policy receives two RGB images from the wrist cameras, Cartesian pose with orientation represented by zyx Euler angles, and Cartesian velocity. It outputs $6$-DoF Cartesian delta poses.
The success-classifier inputs and environment settings remain task-dependent and are therefore reported separately in Tab.~\ref{tab:appendix:task_config}.

\begin{table}[ht!]
    \caption{Shared hardware and runtime details used across all tasks.}
    \label{tab:appendix:shared_setup}
    \centering
    \small
    \begin{tabular}{|l|p{9.5cm}|}
    \hline
    \textbf{Parameter} & \textbf{Value}\\
    \hline
    Robot platform & Franka Emika Robot with Franka Hand\\
    \hline
    Control compute & NVIDIA Jetson Orin AGX with L4T and PREEMPT\_RT kernel\\
    \hline
    Policy / classifier compute & Desktop PC with Ubuntu 22.04 and NVIDIA RTX 4090\\
    \hline
    Policy command rate & $10$~Hz\\
    \hline
    Cameras & Two Intel RealSense D405 wrist cameras\\
    \hline
    Observation space & Two RGB wrist-camera images, Cartesian pose (zyx Euler convention), Cartesian velocity\\
    \hline
    Action space & $6$-DoF Cartesian delta pose\\
    \hline
    Raw camera resolution & $1280 \times 720$\\
    \hline
    Network input resolution & $128 \times 128$\\
    \hline
    Camera exposure & $40{,}000~\mu\mathrm{s}$\\
    \hline
    Gripper usage & Objects are pre-grasped, and regrasping is not required in the reported tasks\\
    \hline
    \end{tabular}
\end{table}

\subsection{Peg Insertion}

This task is taken from the Functional Manipulation Benchmark (FMB) \cite{luo2025fmb}.
The peg and its counterpart are 3D-printed with white material.
The clearance is $\approx 2$ mm on the diameter.
The peg is initially grasped by the robot and retained during the experiment.
The robot fingers are designed such that the peg cannot drop out of its grip, but it can move slightly in the end effector.
The objective is to insert the peg into a matching hole on the benchmark board.
See also Fig.~\ref{fig:setups_full}.

\subsection{RAM Insertion}

This task follows the RAM insertion setting used in the original HIL-SERL work \cite{luo2025precise}.
At the beginning of each episode, the robot has already grasped the RAM stick.
The objective is to align the module with the correct socket and insert it until it is properly seated.
Compared with peg insertion, this task is visually and mechanically more demanding because the insertion tolerances are tighter and partial insertion states are common.
We consider an episode successful once the RAM stick is fully inserted into the designated slot.
We modified the clips at the sides of the slot on the mainboard to enable the robot to remove the RAM stick from its slot by itself.

\subsection{Limit Fixture}

The limit fixture task is taken from an existing high-value, real-world electronics manufacturing workflow.
In this problem, the robot must insert a fixture into a tray that is used in a downstream testing step.
The fixture is necessary for the subsequent step, in which a busbar is inserted.
The task is representative of industrial contact-rich assembly, as successful completion requires accurate alignment and reliable insertion under limited clearances.
Because the benchmark originates from a proprietary production process, we cannot disclose the full geometry or all process-specific details.
\subsection{Busbar}

The busbar task is drawn from the same production workflow as the limit fixture assembly and is the direct follow-up step.
Here, the robot inserts a busbar into its target location after the preceding fixture-related preparation step.
The purpose of the busbar is to conduct electrical power across the tray.
This benchmark is again contact-rich and requires accurate positioning during the final insertion phase.
As with the limit fixture task, some setup details cannot be reported in full due to confidentiality constraints.
\subsection{Task Configuration}

Table~\ref{tab:appendix:task_config} summarizes the episode length, action scales, and reset randomization for the four tasks studied in the paper.

\begin{table*}[ht!]
    \caption{Task-specific configuration for the four RL tasks in the paper.}
    \label{tab:appendix:task_config}
    \centering
    \begin{tabular}{|l|c|p{3.0cm}|c|c|}
    \hline
    \textbf{Task} & \textbf{\shortstack{Episode length\\{}[steps]}} & \textbf{\shortstack{Action scale\\(translation [m],\\rotation [rad])}} & \textbf{\shortstack{$x,y$ reset\\range [m]}} & \textbf{\shortstack{Yaw reset\\range [rad]}} \\
    \hline
    Peg insertion & $150$ & $(0.01, 0.02)$ & $\pm 0.02$ & $\pm 0.05$ \\
    \hline
    RAM insertion &  $200$ & $(0.005, 0.02)$ & $\pm 0.02$ & $\pm 0.05$ \\
    \hline
    Limit fixture &  $150$ & $(0.01, 0.02)$ & $\pm 0.02$ & $\pm 0.05$ \\
    \hline
    Busbar & $150$ & $(0.01, 0.02)$ & $\pm 0.02$ & $\pm 0.05$ \\
    \hline
    \end{tabular}
\end{table*}

\section{Policy and Training Details}\label{app:training}

This section records the reinforcement-learning configuration used across tasks.
The defaults in this section are shared across tasks.

\subsection{Training Defaults}

\begin{table}[ht!]
    \caption{Shared policy and training details.}
    \label{tab:appendix:training}
    \centering
    \small
    \begin{tabular}{|l|p{6cm}|}
    \hline
    \textbf{Parameter} & \textbf{Value}\\
    \hline
    Learning algorithm & RLPD with SAC-style pixel agent\\
    \hline
    Batch size & $256$ total = $128$ demo + $128$ online RL\\
    \hline
    Critic-to-actor ratio & $2$\\
    \hline
    Discount factor $\gamma$ & $0.97$\\
    \hline
    Maximum training steps & $1{,}000{,}000$ gradient steps\\
    \hline
    Replay buffer capacity & $200{,}000$ transitions\\
    \hline
    Random exploration steps & $0$\\
    \hline
    Training starts after & $100$ online transitions\\
    \hline
    Policy publication / update interval & Every $50$ environment steps\\
    \hline
    Evaluation period & Every $2{,}000$ training steps\\
    \hline
    Evaluation rollouts during training & $5$ trajectories\\
    \hline
    Final evaluation protocol & $100$ trials per task\\
    \hline
    Demonstration budget & $20$ demonstrations per method and task\\
    \hline
    \end{tabular}
\end{table}

\subsection{Architecture Details}

Table~\ref{tab:appendix:actor_critic} summarizes the actor, critic, and optimization settings used for policy learning.
Table~\ref{tab:appendix:vision_encoder} reports the vision encoder and image augmentation details.

\begin{table}[ht!]
    \caption{Actor, critic, and optimization details.}
    \label{tab:appendix:actor_critic}
    \centering
    \begin{tabular}{|p{6cm}|p{7cm}|}
    \hline
    \textbf{Parameter} & \textbf{Value}\\
    \hline
    Policy hidden dimensions & $[256, 256]$\\
    \hline
    Policy activation & \texttt{tanh}\\
    \hline
    Policy output distribution & Gaussian with $\tanh$ squashing\\
    \hline
    Standard-deviation parameterization & \texttt{exp}\\
    \hline
    Standard-deviation range & $[10^{-5}, 5]$\\
    \hline
    Policy layer normalization & Enabled\\
    \hline
    Critic hidden dimensions & $[256, 256]$\\
    \hline
    Critic activation & \texttt{tanh}\\
    \hline
    Critic layer normalization & Enabled\\
    \hline
    Critic ensemble size & $2$ (clipped double-$Q$)\\
    \hline
    Temperature initialization & $10^{-2}$\\
    \hline
    Target entropy & negative half of action dimension (automatic)\\
    \hline
    Learning rate & $3 \times 10^{-4}$ for all networks\\
    \hline
    Polyak averaging coefficient $\tau$ & $0.005$\\
    \hline
    Reward bias & $0.0$\\
    \hline
    \end{tabular}
\end{table}

\begin{table}[ht!]
    \caption{Vision encoder and data augmentation details.}
    \label{tab:appendix:vision_encoder}
    \centering
    \begin{tabular}{|p{5.0cm}|p{4.1cm}|}
    \hline
    \textbf{Parameter} & \textbf{Value}\\
    \hline
    Encoder type & Pretrained frozen ResNet-10\\
    \hline
    Image pretraining & ImageNet\\
    \hline
    Pre-pooling & Enabled\\
    \hline
    Pooling method & Spatial learned embeddings\\
    \hline
    Number of spatial blocks & $8$\\
    \hline
    Bottleneck dimension & $256$\\
    \hline
    Network image size & $128 \times 128$\\
    \hline
    Data augmentation & Batched random crop\\
    \hline
    Crop padding & $4$ pixels\\
    \hline
    \end{tabular}
\end{table}

\section{Reward and Success Classification}\label{app:classifier}

The main text describes a sparse binary reward signal, and the implementation matches that description across all four tasks.
Each task uses a task-specific learned visual success classifier whose output is thresholded online to obtain the binary reward used for reinforcement learning.
The classifier input views differ by task, but the reward construction itself is shared: for all four tasks, we use the binary reward $r_t=1$ if $\text{sigmoid}(f_\psi(o_t)) > 0.85$ and $r_t=0$ otherwise.

\subsection{Classifier Details}

Following the reward-design recipe of \cite{luo2025precise}, we use a separate learned binary success classifier for each task.
The classifier maps the current visual observation to a scalar success score $f_\psi(o_t)$, which is then thresholded to obtain the binary reward used during reinforcement learning.
We keep one classifier per task rather than sharing a single model across all tasks, since the visual signatures of success differ substantially between peg insertion, RAM insertion, limit fixture assembly, and busbar assembly.

\begin{table}[ht!]
    \caption{Success-classifier architecture and optimization details.}
    \label{tab:appendix:classifier_details}
    \centering
    \begin{tabular}{|p{5.0cm}|p{5.0cm}|}
    \hline
    \textbf{Parameter} & \textbf{Value}\\
    \hline
    Classifier inputs & Task-relevant wrist-camera images\\
    \hline
    Backbone & Pretrained frozen ResNet-10\\
    \hline
    Pooling & Spatial learned embeddings\\
    \hline
    Bottleneck dimension & $256$\\
    \hline
    Classification head & Two-layer MLP\\
    \hline
    Loss & Cross-entropy\\
    \hline
    Optimizer & Adam\\
    \hline
    Learning rate & $3 \times 10^{-4}$\\
    \hline
    Optimization iterations & $100$\\
    \hline
    Reward threshold & $\text{sigmoid}(f_\psi(o_t)) > 0.85$\\
    \hline
    \end{tabular}
\end{table}

\subsection{Classifier Dataset and Hard Negatives}

The classifier dataset is collected before policy training by recording short teleoperated rollouts and storing the corresponding images together with binary success labels.
During rollout, while the user is pressing a button, all frames are labelled as positives, while all remaining frames are treated as negatives.
For each task, the dataset contains approximately $3{,}000$ negative and $1{,}500$ positive samples.
To make the classifier robust, we deliberately include difficult negative examples that are visually similar to success, for example states with incomplete insertion, asymmetric seating, contact at the wrong slot (in case of RAM insertion), or other near-goal misalignments.
These hard negatives are important because they reduce false positives during online reward evaluation and make the sparse reward considerably more reliable.
After training, the classifier is frozen and used only for reward computation, and it is not optimized jointly with the policy during RL.

\section{Controller Stack Parameters}\label{app:controller_params}

This section records the controller, guide-mode, and reference generator parameters used in the experimental comparison.
To distinguish the HIL-SERL baseline from our controller stack, Tab.~\ref{tab:appendix:controller_stack} contrasts the two impedance-layer parameterizations: the baseline uses a manually specified Cartesian damping matrix together with asymmetric error clipping, whereas our stack uses damping factors and no clipping.
The damping factors are used during double diagonalization to calculate a damping matrix in modal space which is then transformed back into Cartesian space.
The velocity/acceleration limits of our stack depend on the task-specific action scales listed in Tab.~\ref{tab:appendix:task_config}.
The clipping values are physical constraints rather than free controller gains. We retain the baseline stiffness and compute impedance damping from inertia and stiffness using six damping factors $\zeta_i=0.7$. Admittance tuning selects low positive-definite $M_a$, increases positive-definite $D_a$ until interaction is stable, and sets $K_a=\boldsymbol{0}$. We reuse this setting across tasks. The reference pole $\omega=30$~s$^{-1}$ is near the Nyquist angular frequency of the $10$-Hz action stream and far below the $1$-kHz torque loop. This reuse is not a sensitivity study or stability proof.

\begin{table}[ht!]
    \caption{Baseline-versus-ours comparison of the controller parameterization.}
    \label{tab:appendix:controller_stack}
    \centering
    \begin{tabular}{|p{5.0cm}|p{3.7cm}|p{3.7cm}|}
    \hline
    \textbf{Parameter} & \textbf{Baseline} & \textbf{Ours}\\
    \hline
    Cartesian translational stiffness & $\left[3000,3000,3000\right]$~N/m & $\left[3000,3000,3000\right]$~N/m \\
    \hline
    Cartesian rotational stiffness & $\left[300,300,300\right]$~N\,m/rad & $\left[300,300,300\right]$~N\,m/rad \\
    \hline
    Cartesian translational damping & $\left[89,89,89\right]$~N\,s/m& N/A \\
    \hline
    Cartesian rotational damping & $\left[7,7,7\right]$~N\,m\,s/rad & N/A \\
    \hline
    Cartesian translational damping factors & N/A & $\left[0.7,0.7,0.7\right]$ \\
    \hline
    Cartesian rotational damping factors & N/A & $\left[0.7,0.7,0.7\right]$ \\
    \hline
    Translational clip $(+x,+y,+z)$ & $(7.5, 1.6, 5.5)\times 10^{-3}$ & N/A \\
    \hline
    Translational clip $(-x,-y,-z)$ & $(2.0, 1.6, 5.0)\times 10^{-3}$ & N/A \\
    \hline
    Rotational clip $(+x,+y,+z)$ & $(10, 25, 5)\times 10^{-3}$ & N/A \\
    \hline
    Rotational clip $(-x,-y,-z)$ & $(10, 25, 5)\times 10^{-3}$ & N/A \\
    \hline
    Transl./rot. integral gains & $0$ / $0$ & $0$ / $0$ \\
    \hline
    Admittance mass translation $M_a$ & N/A & $\left[1,1,1\right]$~kg \\
    \hline
    Admittance mass rotation $M_a$ & N/A & $\left[0.1,0.1,0.1\right]$~kg\,m$^2$ \\
    \hline
    Admittance damping translation $D_a$ & N/A & $\left[100,100,100\right]$~N\,s/m  \\
    \hline
    Admittance damping rotation $D_a$ & N/A & $\left[20,20,20\right]$~N\,m\,s/rad  \\
    \hline
    Admittance stiffness $K_a$ & N/A & $\mathbf{0}_{6\times 6}$ \\
    \hline
    Reference generator bandwidth $\omega$ & N/A & $30$~s$^{-1}$ \\
    \hline
    \end{tabular}
\end{table}

\begin{table*}[ht!]
    \caption{Task-specific reference-generator limits of our stack derived from the action scales in Tab.~\ref{tab:appendix:task_config}. The $[\min,\max]$ entries are signed directional bounds.}
    \label{tab:appendix:reference_limits}
    \centering
    \small
    \resizebox{\textwidth}{!}{%
    \begin{tabular}{|l|c|c|c|c|c|c|}
    \hline
    \textbf{Task}  & \textbf{Trans. vel.} & \textbf{Trans. acc.} & \textbf{Rot. vel.} & \textbf{Rot. acc.} & \textbf{Trans. jerk} & \textbf{Rot. jerk}\\
     &  \textbf{$[\min,\max]$} & \textbf{$[\min,\max]$} & \textbf{$[\min,\max]$} & \textbf{$[\min,\max]$} & \textbf{$\max$} & \textbf{$\max$}\\
     &  \textbf{[m/s]} & \textbf{[m/s$^2$]} & \textbf{[rad/s]} & \textbf{[rad/s$^2$]} & \textbf{[m/s$^3$]} & \textbf{[rad/s$^3$]}\\
    \hline
    Peg insertion & $[-0.09, 0.10]$ & $[-0.90, 1.00]$ & $[-0.18, 0.20]$ & $[-1.8, 2.0]$ & $10000.0$ & $10000.0$\\
    \hline
    RAM insertion & $[-0.045, 0.05]$ & $[-0.45, 0.50]$ & $[-0.18, 0.20]$ & $[-1.8, 2.0]$ & $10000.0$ & $10000.0$\\
    \hline
    Limit fixture & $[-0.09, 0.10]$ & $[-0.90, 1.00]$ & $[-0.18, 0.20]$ & $[-1.8, 2.0]$ & $10000.0$ & $10000.0$\\
    \hline
    Busbar & $[-0.09, 0.10]$ & $[-0.90, 1.00]$ & $[-0.18, 0.20]$ & $[-1.8, 2.0]$ & $10000.0$ & $10000.0$ \\
    \hline
    \end{tabular}}
\end{table*}

\section{Demonstration and Intervention Protocol}\label{app:demo}

This section provides the operator-side data-collection protocol used for demonstrations and corrective interventions.
We report it separately because these workflow details strongly affect real-world training time and usability, even though they are not part of the core control or learning equations.
Unless otherwise stated, the same demonstration and intervention logic is used across all four tasks.

\subsection{Demonstration Collection}

Demonstrations are collected either with a SpaceMouse or through kinesthetic guidance.
In both cases, the object is initially held in the gripper, and the reported tasks do not require regrasping.
For each method and task, we collect $20$ demonstrations before RL training.
The operator then moves the robot to the goal pose, either by using the SpaceMouse or, in the kinesthetic setting, by guiding the robot by hand while guide mode remains active throughout the demonstration.
A demonstration ends when the task succeeds or reaches the maximum episode length. The robot then moves upward and returns to its reset pose.
The reset pose is randomized in the $x$ and $y$ directions and around the $z$ axis before the next episode, with task-specific magnitudes listed in Tab.~\ref{tab:appendix:task_config}.
One trained, unblinded operator performed all demonstrations and interventions. During kinesthetic demonstrations, the operator guided the robot from above with both hands outside the wrist-camera views. The operator practiced both interfaces until completing the tasks reliably. Practice rollouts were excluded from the reported times and replay buffers.

\subsection{Human Interventions During Training}

During RL training, the operator intervenes when the policy becomes unsafe, loses alignment, or behaves suboptimally. Intervention overrides the policy with SpaceMouse or kinesthetic input according to the evaluated condition. The resulting observation-action-reward tuples are retained, and the corrective trajectory is added to both the demonstration and online RL replay buffers so it can affect subsequent updates.

\end{document}